\documentclass[]{aidata}

\usepackage[toc,page,header]{appendix}
\usepackage{minitoc}
\usepackage{cleveref} 
\usepackage{subcaption}
\usepackage{booktabs}

\usepackage[table,dvipsnames]{xcolor}
\usepackage{array}
\usepackage{enumitem}
\usepackage{tabularx}
\usepackage{ragged2e}
\usepackage[ruled,vlined,linesnumbered]{algorithm2e}
\usepackage{multirow}

\definecolor{myblue}{RGB}{232, 240, 254}
\definecolor{mypurple}{RGB}{243, 229, 245}
\definecolor{mygreen}{RGB}{232, 245, 233}
\definecolor{lightblue}{RGB}{230,240,255}
\definecolor{lightpink}{RGB}{255,230,240}
\definecolor{lightyellow}{RGB}{255,250,230}
\definecolor{lightgreen}{RGB}{230,255,230}
\definecolor{lightpurple}{RGB}{240,230,255}
\definecolor{reflectbg}{RGB}{245,245,250}
\definecolor{repibg}{RGB}{245,250,245}
\definecolor{qualifybg}{RGB}{255,250,240}

\newcommand{\coloredblock}[2]{\colorbox{#1}{\parbox{\dimexpr\linewidth-2\fboxsep}{#2}}}

\newcolumntype{C}{>{\Centering\arraybackslash}X}

\usepackage{amsmath,amssymb,mathtools,amsthm}

\theoremstyle{plain}

\theoremstyle{definition}

\theoremstyle{remark}

\usepackage[most]{tcolorbox}

\newtcbox{\hlprimarytab}{on line, rounded corners, box align=base,
  colback=green!10, colframe=white, size=fbox, arc=3pt,
  before upper=\strut, top=-2pt, bottom=-4pt, left=-2pt, right=-2pt, boxrule=0pt}

\newtcbox{\hlsecondarytab}{on line, rounded corners, box align=base,
  colback=red!10, colframe=white, size=fbox, arc=3pt,
  before upper=\strut, top=-2pt, bottom=-4pt, left=-2pt, right=-2pt, boxrule=0pt}

\newcommand{\dashifted}{\raisebox{0.5\depth}{\tiny$\downarrow$}}
\newcommand{\uashifted}{\raisebox{0.5\depth}{\tiny$\uparrow$}}
\newcommand{\dar}[1]{{\raisebox{0.6ex}{\tiny\hlsecondarytab{\dashifted\,#1}}}}
\newcommand{\uar}[1]{{\raisebox{0.6ex}{\tiny\hlprimarytab{\uashifted\,#1}}}}

\usepackage{listings}
\tcbuselibrary{skins, breakable}

\title{Socratic-Geo: Synthetic Data Generation and Geometric Reasoning via Multi-Agent Interaction}

\author[1,2,3,*]{Zhengbo Jiao}
\author[2,*]{Shaobo Wang}
\author[1,4,*]{Zifan Zhang}          
\author[1]{Wei Wang}
\author[1,\textsuperscript{$\blacktriangle$}]{Bing Zhao}  
\author[1,\dagger]{Hu Wei}
\author[2,\dagger]{Linfeng Zhang}

\affiliation[1]{AI DATA, Alibaba Group Holding Limited}
\affiliation[2]{EPIC Lab, Shanghai Jiao Tong University}
\affiliation[3]{Shanghai University of Finance and Economics}
\affiliation[4]{Wuhan University}

\contribution[*]{Equal contribution}

\contribution[\textsuperscript{$\blacktriangle$}]{Project leader}
\contribution[\dagger]{Corresponding authors}

\abstract{
Multimodal Large Language Models (MLLMs) have significantly advanced vision-language understanding. However, even state-of-the-art models struggle with geometric reasoning, revealing a critical bottleneck: the extreme scarcity of high-quality image-text pairs. Human annotation is prohibitively expensive, while automated methods fail to ensure fidelity and training effectiveness. Existing approaches either passively adapt to available images or employ inefficient random exploration with filtering, decoupling generation from learning needs. We propose Socratic-Geo, a fully autonomous framework that dynamically couples data synthesis with model learning through multi-agent interaction. The Teacher agent generates parameterized Python scripts with reflective feedback (Reflect for solvability, RePI for visual validity), ensuring image-text pair purity. The Solver agent optimizes reasoning through preference learning, with failure paths guiding Teacher's targeted augmentation. Independently, the Generator learns image generation capabilities on accumulated "image-code-instruction" triplets, distilling programmatic drawing intelligence into visual generation. Starting from only 108 seed problems, Socratic-Solver achieves 49.11 on six benchmarks using one-quarter of baseline data, surpassing strong baselines by 2.43 points. Socratic-Generator achieves 42.4\% on GenExam, establishing new state-of-the-art for open-source models, surpassing Seedream-4.0 (39.8\%) and approaching Gemini-2.5-Flash-Image (43.1\%).
}

\date{\today}
\correspondence{\email{zhanglinfeng1997@outlook.com}, \email{kongwang@alibaba-inc.com}}

\checkdata[Project Page]{\url{https://github.com/Frostlinx/Socratic-geo}}

\begin{document}

\maketitle

\begin{figure*}[tb!]
    \centering
    \includegraphics[width=1\linewidth]{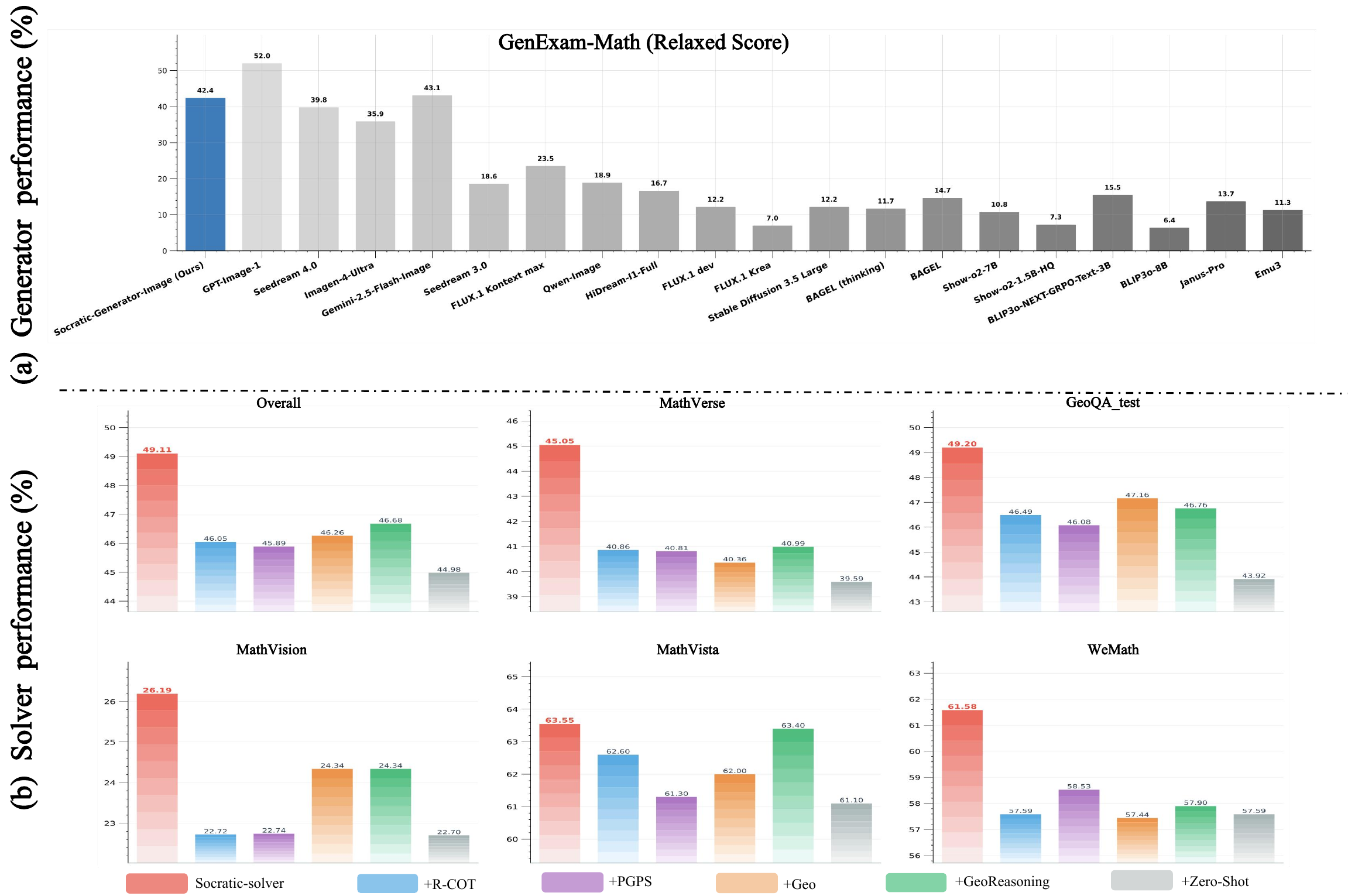}
    \caption{Overall performance comparison demonstrating the dual effectiveness of the Socratic-Geo framework in both reasoning and generation. (a) Our Socratic-Generator-Image achieves an impressive 42.4 Relaxed score on the \textit{GenExam-Math} benchmark, establishing a new state-of-the-art for open-source models and matches strong closed-source systems like Gemini-2.5-Flash-Image. (b) Our Socratic-Solver achieves an impressive 49.11\% average accuracy across the reasoning benchmarks, marking a substantial +4.13 point improvement over the zero-shot baseline and consistently outperforming all other fine-tuning methods.}
    \label{fig:result}
    \vspace{-1em}
\end{figure*}

\section{Introduction}

The burgeoning development of Multimodal Large Models (MLLMs) has significantly advanced the intersection of vision and language. As foundational capabilities mature, multimodal mathematical reasoning has garnered significant research attention~\citep{zhang2024mathversedoesmultimodalllm, lu2024mathvistaevaluatingmathematicalreasoning, chen2022geoqageometricquestionanswering}, especially in geometry—a domain that demands both visual perception and logical deduction. However, progress is severely bottlenecked by the extreme scarcity of high-quality geometric training data. Manual annotation is prohibitively expensive and time-consuming, while automated synthesis struggles to simultaneously balance correctness, diversity, and training effectiveness. This raises a critical question: \textit{How can we design an efficient geometric data synthesis engine?}

Existing automated methods fall into three categories. First, image-based textual augmentation~\citep{deng2024r, xin2025generalizablegeometricimagecaption} enhances linguistic quality but remains \textit{passive}—refining descriptions for pre-existing images without constructing new structures. Second, symbolic-driven random generation~\citep{fu2025trustgeogenformalverifieddataengine, lu2021intergpsinterpretablegeometryproblem} ensures correctness through formal languages but adopts \textit{blind exploration}, generating vast candidates then filtering heuristically. Third, LLM-driven augmentation~\citep{zhang2024mavismathematicalvisualinstruction} produces diverse content but acts as \textit{black-box amplifiers}, inheriting model biases and lacking fine-grained control. Most critically, all paradigms produce \textit{static, one-way} datasets that decouple synthesis from learning—data generation occurs independently of model training, missing opportunities for iterative improvement.

To address these challenges, we propose Socratic-Geo: a \textit{dynamic} data synthesis engine that couples generation with learning (Figure \ref{fig:socratic}). Inspired by the Socratic method, our framework implements goal-driven synthesis through multi-agent interaction. The Teacher agent generates parametric Python scripts using a "conceive-and-verify" loop where Reflect (solvability checker) ensures mathematical validity while RePI (visual validator) verifies rendering correctness, enabling \textit{proactive structural modification} rather than passive adaptation. For targeted enhancement, Solver's failed reasoning paths diagnose weaknesses and guide Teacher's generation goals, replacing blind exploration with \textit{learner-driven synthesis}. Finally, we fine-tune the Generator on accumulated "image-code-instruction" triplets, distilling programmatic drawing intelligence into generative capabilities and establishing a \textit{synthesis-learning closed loop}. Our main contributions are threefold:

\begin{itemize}[leftmargin=10pt, topsep=0pt, itemsep=1pt, partopsep=1pt, parsep=1pt]

\item \textbf{Goal-Driven Programmatic Synthesis.} We introduce a paradigm that establishes generation goals by diagnosing learner weaknesses and enhances image structures through code modification, supported by our Reflect-RePI mechanism for proactive self-correction.

\item \textbf{Multi-Agent Interaction Framework.} We propose Socratic-Geo that tightly couples synthesis with learning. Teacher-Solver interaction drives reasoning evolution, extending to Generator for controllable geometric image generation from programmatic data.

\item \textbf{Strong Empirical Results.} Starting from only 108 seed problems, Socratic-Solver achieves substantial improvements across benchmarks, with an overall +4.13 point gain over baselines using merely a quarter of training samples (Figure~\ref{fig:result}b). Independently, Socratic-Generator achieves 42.4\% on GenExam (Figure~\ref{fig:result}a), establishing new state-of-the-art for open-source models by surpassing commercial model Seedream-4.0 (39.8\%) and approaching Gemini-2.5-Flash-Image (43.1\%).

\end{itemize}

\section{Related Work}

\subsection{Geometry Generation}
To address data scarcity, several methods have been proposed for synthetic geometric problem generation. 
Inter-GPS~\citep{lu2021intergpsinterpretablegeometryproblem} parses diagrams into formal logic to enable symbolic reasoning but relies on manually drawn images. 
G-LLaVA~\citep{gao2025gllavasolvinggeometricproblem} constructs Geo170K, a large-scale dataset of geometry problems paired with images. 
R-CoT~\citep{deng2024r} introduces reverse chain-of-thought to generate logically consistent question–answer pairs from diagrams. 
TrustGeoGen~\citep{fu2025trustgeogenformalverifieddataengine} employs formal verification to ensure generated problems are mathematically valid. 
Generalizable Geometric Image Caption Synthesis~\citep{xin2025generalizablegeometricimagecaption} focuses on improving caption diversity and generalization.  
While these approaches improve data scale and logical consistency, they typically rely on templates, offering limited control over geometric structure.

\subsection{Multi-agent Interaction}
Multi-agent frameworks have emerged as a means to generate training data autonomously.  
LLM2LLM~\citep{lee2024llm2llmboostingllmsnovel} uses iterative refinement between teacher and student models.  
R-Zero~\citep{huang2025rzeroselfevolvingreasoningllm} and Absolute Zero~\citep{zhao2025absolutezeroreinforcedselfplay} achieve self-evolution without human data through reinforcement learning and self-play.  
Socratic-Zero~\citep{wang2025socraticzerobootstrappingreasoning} implements co-evolution among Teacher, Solver, and Generator agents to produce reasoning data.  
Vision-Zero~\citep{wang2025visionzeroscalablevlmselfimprovement} extends this idea to vision–language models via gamified self-play.  
SPICE~\citep{liu2025spiceselfplaycorpusenvironments} improves reasoning through self-play within corpus environments.  
These methods demonstrate strong capabilities in text-based reasoning tasks, but similar collaborative mechanisms have not yet been applied to visual reasoning tasks involving structured diagram generation.

\section{Preliminaries}

\subsection{Group Relative Policy Optimization}

Group Relative Policy Optimization (GRPO)~\citep{shao2024deepseekmathpushinglimitsmathematical} is a policy-gradient algorithm designed for reinforcement learning tasks where model outputs can be scored via deterministic verification rules.
Given an input problem $q$, the reference policy $\pi_{\text{ref}}$ generates $G$ candidate outputs $\{o_1, \dots, o_G\}$, each assigned a scalar reward $R_i$ by a verifiable evaluation function. The sequence-level advantage is computed via group normalisation:
\begin{equation}
A^{(i)} = \frac{R_i - \text{mean}(\{R_j\}_{j=1}^G)}
               {\text{std}(\{R_j\}_{j=1}^G)},
\end{equation}
where $\text{mean}$ and $\text{std}$ denote the sample mean and standard deviation of the group's rewards.

The optimisation objective applies PPO's clipped surrogate at the token level, with KL regularisation to constrain policy deviation:
\begin{equation}
\begin{aligned}
\mathcal{L}_{\mathrm{GRPO}}(\theta) &=
\frac{1}{G} \sum_{i=1}^G 
\frac{1}{|o_i|} \sum_{t=1}^{|o_i|}
\min\big(r_t(\theta)A^{(i)},~ \mathrm{clip}(r_t(\theta), \\
& \qquad 1-\epsilon,\,1+\epsilon)A^{(i)}\big)
 - \beta\,D_{\mathrm{KL}}\!\left(\pi_{\theta} \,\|\, \pi_{\mathrm{ref}}\right)
\end{aligned}
\end{equation}

where 
$r_t(\theta) = \frac{\pi_{\theta}(o_{i,t} \mid q)}{\pi_{\text{ref}}(o_{i,t} \mid q)}$,  
$\epsilon$ is the clipping range, and $\beta$ controls the KL regularisation strength.

By leveraging group-relative advantages from verifiable rewards, GRPO streamlines training by removing the value network, reducing computational overhead, and stabilising learning in sparse-reward environments.

\subsection{Socratic-Zero Framework and Limitations}
Socratic-Zero is a multi-agent learning framework for text-based mathematical reasoning, comprising a \textit{Solver}, \textit{Teacher}, and \textit{Generator}. The \textit{Teacher} refines the \textit{Solver}'s reasoning by generating improved text problems, and the \textit{Generator} learns this text-to-text transformation to expand the dataset.

This approach is incompatible with geometry, where diagrams are integral to problem definitions and must align precisely with textual statements. In geometry, correctness depends on axiomatic consistency between text and figure. The \textit{Teacher}, limited to language, cannot use tools to programmatically modify diagrams (e.g., add auxiliary lines) or verify geometric integrity, leading to unreliable (image, text, solution) triplets.

Without such structured, tool-based outputs, the \textit{Generator} learns only from vague textual prompts to unstructured pixel data, lacking precise supervision. Thus, the original framework cannot synthesize valid reasoning data or high-fidelity geometric images, motivating a paradigm grounded in programmatic control.

\begin{itemize}[leftmargin=12pt]
    \item \textbf{Verify}: Formally compare the \textit{Solver}’s responses with the reference solution to detect incorrect reasoning cases requiring intervention.
    \item \textbf{Analyze}: Perform dual-modality error diagnosis. For the geometric \textbf{code} (and its rendered diagram), inspect structural properties and detect missing or violated axioms; for the \textbf{text} (problem statement), identify semantic inconsistencies, underspecified constraints, or misleading descriptions. This step pinpoints the minimal modifications needed to repair or strengthen the problem.
    \item \textbf{Invent} (via RePI): Programmatically modify the underlying Python geometry code to construct a new problem that explicitly incorporates the critical constraints found in the analysis. Ensure the modified code executes successfully and that the generated diagram is perfectly aligned with the updated textual description.
    \item \textbf{Qualify} (via Reflect): As a self-verifying step, the \textit{Teacher} solves the invented problem using its reasoning pipeline. Only if the solution matches the reference and the geometry passes all checks is the example admitted to the reasoning curriculum and used to generate paired (text, image) data for the \textit{Generator}.
\end{itemize}

\section{Method}
\label{sec:method}


\begin{figure*}[tb!]
    \centering
    \includegraphics[width=1\linewidth]{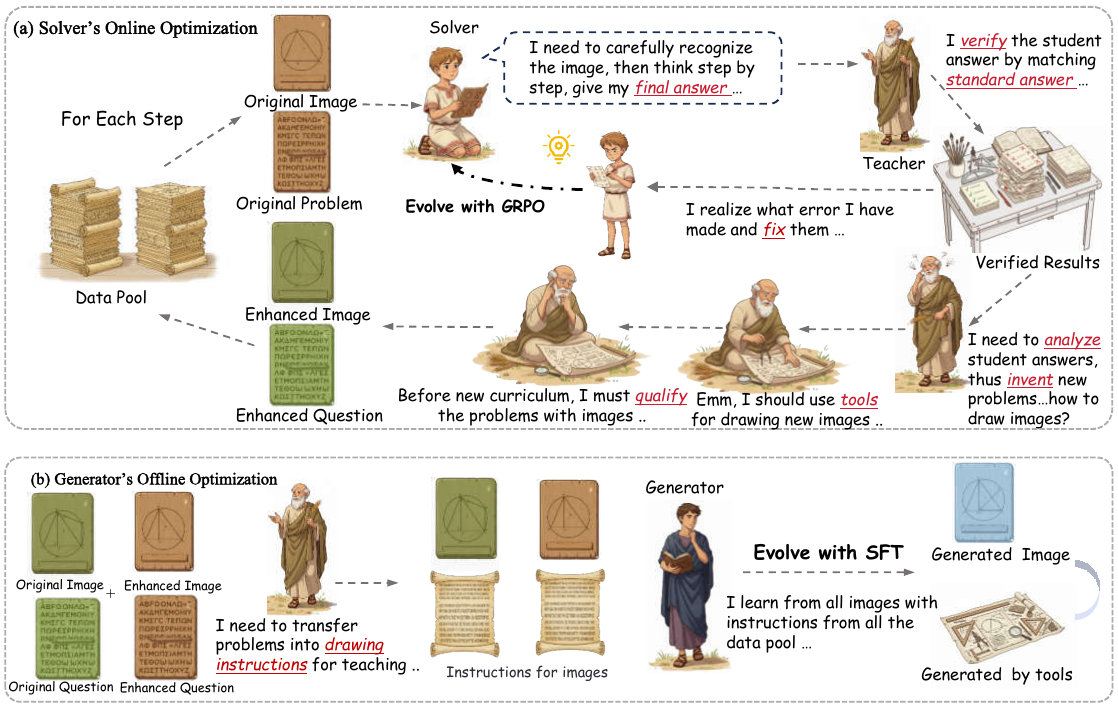}
    \caption{Overview of the Socratic-Geo framework. (a) A closed-loop reasoning process where the Solver’s failures trigger the Teacher to invent and validate new problems, enriching the curriculum. (b) The Generator distills the Teacher’s programmatic drawing instructions into a standalone image synthesis model.}
    \label{fig:socratic}
\end{figure*}

\subsection{Framework Overview}
We propose \textbf{Socratic-Geo}, a fully autonomous framework designed for \textit{goal-driven data synthesis} in the geometric domain. Operating from a minimal seed set without reliance on external data, the framework functions as a closed-loop synthesis engine. Its \textit{primary objective} is to continuously generate challenging (\textit{image}, \textit{text}, \textit{solution}) triplets for reasoning model training. As a \textit{synergistic secondary objective}, the framework leverages valuable assets created during this process to train a high-fidelity image \textit{Generator}. This entire synthesis process is driven by the interaction of three specialized agents:

As illustrated in Figure~\ref{fig:socratic}, these agents form an interaction loop centered on reasoning. The \textit{Solver}'s struggles trigger the \textit{Teacher}'s invention process. The \textit{Teacher}'s programmatic invention, in turn, provides gold-standard data for the \textit{Solver}'s next learning stage. This core \textit{Solver-Teacher} interaction ensures that the synthesized curriculum continuously adapts to the \textit{Solver}'s evolving capabilities. The curriculum, $\mathcal{C}$, evolves iteratively as:
\begin{equation}
\mathcal{C}_{t+1} = \mathcal{C}_t \cup \{ (I', q', a') \mid \exists (I, q, a^*) \in \mathcal{C}_t,~ \{a_{\mathcal{S}}^{(i)}\}_{i=1}^k \sim \pi_{\mathcal{S}}^{(t)}(I,q),~ \sum_{i=1}^k \mathcal{V}(q, a_{\mathcal{S}}^{(i)}, a^*) = 0 \}
\label{eq:curriculum-update}
\end{equation}
where problems from the current curriculum $\mathcal{C}_t$ that the \textit{Solver} fails to solve across $k$ attempts trigger the \textit{Teacher}'s invention pipeline, producing a new validated triplet that is appended to form $\mathcal{C}_{t+1}$.

\subsection{Teacher Engine: Core of Data Synthesis}
The \textit{Teacher} acts as a proactive constructor of geometric reasoning data through a pipeline (Fig.~\ref{fig:socratic}):

Through this process, each synthesized (image, text, solution) triplet is executable, geometrically sound, textually precise, and targeted to address weaknesses in the \textit{Solver}.

\begin{algorithm}[tb!]
\SetAlgoLined
\DontPrintSemicolon
\scriptsize
\caption{Teacher Modules}

\KwIn{$\mathcal{P}_0$, $\pi_S$, $\pi_T$, $G_\phi$, $T$}
\KwOut{$\mathcal{C}$, $\pi_S^*$, $G_\phi^*$}

$\mathcal{C} \leftarrow \mathcal{P}_0$, $\mathcal{D} \leftarrow \emptyset$\;

\For{$t = 1$ \KwTo $T$}{
    \tcp{Solver Phase}
    \coloredblock{lightblue}{%
        $(I, q, a^*, c) \sim \mathcal{C}$\\
        $a_S \leftarrow \pi_S(I, q)$
    }
    
    \If{$a_S \neq a^*$}{
        \tcp{RePI: Invent}
        \coloredblock{lightpink}{%
            $C \leftarrow \text{Diagnose}(\pi_S, a^*)$
        }
        
        $\text{ok} \leftarrow \text{False}$\;
        \While{not ok}{
            \coloredblock{lightyellow}{%
                $c' \leftarrow \pi_T.\text{Modify}(c, C)$\\
                $(I', q', a') \leftarrow \text{Exec}(c')$\\
                \If{success}{$\text{ok} \leftarrow \text{True}$}
            }
        }
        
        \tcp{Reflect: Verify}
        \coloredblock{lightgreen}{%
            $a_T \leftarrow \pi_T(I', q')$\\
            $v \leftarrow \text{Check}(a_T, a')$
        }
        
        \If{$v$}{
            \coloredblock{lightgreen}{%
                $\mathcal{C} \leftarrow \mathcal{C} \cup \{(I', q', a', c')\}$
            }
            
            \tcp{Generator: Learn}
            \coloredblock{lightpurple}{%
                $d \leftarrow \pi_T.\text{Trans}(q')$\\
                $\mathcal{D} \leftarrow \mathcal{D} \cup \{(d, I')\}$
            }
        }
    }
    
    \colorbox{lightblue}{$\pi_S \leftarrow \text{Train}(\pi_S, \mathcal{C})$}\;
    \colorbox{lightpurple}{$G_\phi \leftarrow \text{Diff}(G_\phi, \mathcal{D})$}\;
}

\Return{$\mathcal{C}$, $\pi_S^*$, $G_\phi^*$}

\end{algorithm}
\subsection{Reinforcement Learning based Solver\protect\\optimization}

The Solver ($\mathcal{S}$) evolves through Group Relative Policy Optimization (GRPO), learning from the high-quality, targeted data supplied by the Teacher. \textbf{Crucially, this is a pure reinforcement learning paradigm rather than knowledge distillation}: the Solver \textit{never} observes the Teacher's chain-of-thought reasoning or reference solutions during training. Instead, it learns exclusively through trial-and-error, receiving only binary reward signals that indicate correctness without revealing the solution path. This design ensures that performance gains stem from the RL mechanism itself, not from imitating teacher outputs.

The Solver and Teacher form the framework's core reasoning loop. For each problem, the Solver with policy $\pi_{\mathcal{S}}$ generates $k$ solution attempts $\{a_{\mathcal{S}}^{(i)}\}_{i=1}^k$. The Teacher provides a binary reward $R_i = \mathcal{V}(q, a_{\mathcal{S}}^{(i)}, a_{\text{ref}}) \in \{0,1\}$ for each attempt.

Its most critical contribution occurs when all attempts fail ($\sum R_i = 0$). In this scenario, the \textit{Teacher}'s own verified reference solution, $a_{\text{ref}}$, is injected as the sole positive example. This ensures the \textit{Solver} receives a gold-standard signal even in complete failure. The positive and negative sets for optimization are thus defined as:
\begin{equation}
(\mathcal{Z}^+, \mathcal{Z}^-) = \begin{cases} (\{a_{\mathcal{S}}^{(i)} \mid R_i = 1\}, \{a_{\mathcal{S}}^{(i)} \mid R_i = 0\}) & \text{if } \sum_{i=1}^k R_i > 0 \\ (\{a_{\text{ref}}\}, \{a_{\mathcal{S}}^{(1)},\ldots,a_{\mathcal{S}}^{(k)}\}) & \text{if } \sum_{i=1}^k R_i = 0 \end{cases}
\label{eq:grpo-sets}
\end{equation}
The Solver parameters $\theta_{\mathcal{S}}$ are then updated to maximize the GRPO objective:

\begin{equation}
\mathcal{L}_{\mathrm{GRPO}}(\theta_{\mathcal{S}}) = - \mathbb{E}_{(I,q) \sim \mathcal{C}_t} \left[ \frac{1}{|\mathcal{Z}|} \sum_{a \in \mathcal{Z}} \frac{1}{|a|} \sum_{t=1}^{|a|} (\hat{A}_t(a) + \beta (r_t - 1)) \log \pi_{\theta_{\mathcal{S}}}(a_t | I, q, a_{<t}) \right]
\label{eq:grpo-loss}
\end{equation}

\begin{figure*}[tb!]
    \centering
    \includegraphics[width=1 \textwidth]{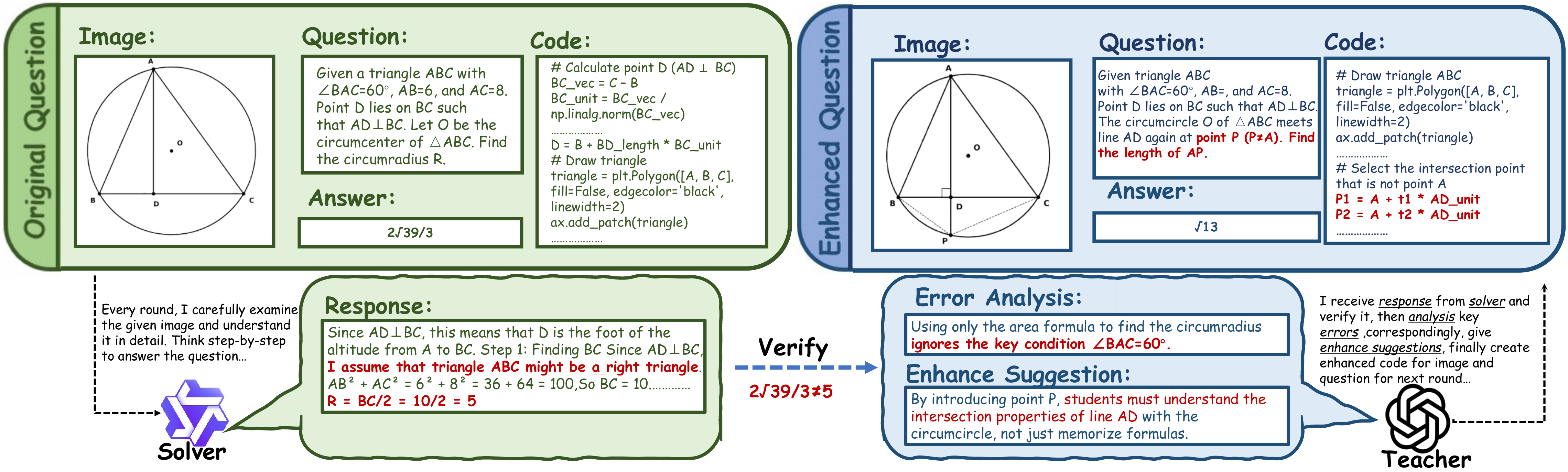}
    \caption{
    A concrete instantiation of the Socratic-Geo interaction pipeline, where the Teacher transforms flawed reasoning into diagnostic problems. Red highlighted regions mark critical intervention points including error diagnosis and geometric augmentation.\\
    \textbf{Left:} The Solver incorrectly assumes a right triangle structure, overlooking the given constraint $\angle BAC = 60^\circ$, leading to an invalid solution path.
    \textbf{Right:} The Teacher introduces point $P$, forcing the Solver to apply the inscribed angle theorem combined with the $60^\circ$ angle property to establish the relationship between $\angle APD$ and $\angle ACD$, directly targeting the reasoning deficiency.
    }
    \vspace{-1em}
    \label{fig:pipeline_instantiation}
\end{figure*}

\subsection{Generator Training: A Synergistic Byproduct}
Crucially, the \textit{Generator} ($\mathcal{G}$) operates independently of the core reasoning loop. It does not interact with the \textit{Solver} or influence the \textit{Teacher}'s decisions. Instead, it is a \textbf{synergistic byproduct} whose training relies entirely on the valuable assets created by the \textit{Teacher} during its primary mission of improving the \textit{Solver}.

To enable this, the \textit{Teacher} performs an \textit{additional, dedicated step} for the \textit{Generator}'s benefit. After inventing a new problem and its programmatic representation, the \textit{Teacher} translates this information into a descriptive, natural-language drawing instruction, $p_{\text{diagram}}$. This creates high-quality (\textit{instruction}, \textit{image}) pairs.

The \textit{Generator}, a diffusion-based model, is then trained on these $(p_{\text{diagram}}, I_{\text{new}})$ pairs. This process is a form of \textit{knowledge distillation}: the symbolic, rule-based, and precise drawing intelligence of the \textit{Teacher} is distilled into the neural weights of the \textit{Generator}. By learning from these programmatic blueprints instead of vague prompts, the \textit{Generator} learns to map structured instructions to geometrically precise diagrams. The supervised fine-tuning (SFT) loss is:
\begin{equation}
\mathcal{L}_{\mathrm{SFT}}(\theta_{\mathcal{G}}) = \mathbb{E}_{p_{\text{diagram}}, z_0, \epsilon, t} \left[ \| \epsilon - \epsilon_{\theta_{\mathcal{G}}}(z_t, t, p_{\text{diagram}}) \|^2 \right], \quad z_t = \sqrt{\bar{\alpha}_t} z_0 + \sqrt{1-\bar{\alpha}_t}\epsilon
\label{eq:sft-loss}
\end{equation}
Here, $\mathcal{E}$ is the VAE encoder, and $\epsilon_{\theta_{\mathcal{G}}}$ is the noise prediction network. This allows us to create a powerful generative model as a valuable byproduct of the main reasoning-focused loop.

\section{Experiments}
\label{sec:formatting}

\subsection{Experiment Setup}

\noindent \textbf{Models} 
We employed Qwen3-VL-235B-A22B-Instruct as the Teacher to produce high-quality question generation and image-code refinement strategies. The Generator component adopted Qwen-Image~\citep{wu2025qwenimage}, distilled from the Teacher's image-generation policy. On the mathematics track of the GenExam text-to-image benchmark, we evaluated Socratic-Geo alongside representative models (see Appendix~\ref{app:models} for complete list). In addition, we trained the Solver on Qwen2.5-VL-7B-Instruct~\citep{bai2025qwen2.5-vl} using GRPO~\citep{shao2024deepseekmathpushinglimitsmathematical} reinforcement learning on both the Socratic-Geo-generated geometry dataset and mainstream geometry datasets.

\noindent \textbf{Benchmarks}  
We evaluated geometric reasoning in mathematics across six benchmarks: MathVerse~\citep{zhang2024mathversedoesmultimodalllm}, MathVista~\citep{lu2024mathvistaevaluatingmathematicalreasoning}, MathVision~\citep{wang2024measuringmultimodalmathematicalreasoning}, GeoQA~\citep{chen2022geoqageometricquestionanswering}, GeomVerse~\citep{kazemi2023geomversesystematicevaluationlarge}, and WeMath~\citep{qiao2024wemathdoeslargemultimodal}. Furthermore, we assessed generator performance on the math track of the GenExam\citep{wang2025genexammultidisciplinarytexttoimageexam}. benchmark.

\begin{table*}[tb!]
\centering
\small
\caption{Evaluation of geometric data synthesis methods on six multimodal reasoning benchmarks: MathVerse, GeomVerse, GeoQA, MathVision, MathVista, and WeMath. Performance is reported in Mean@1 (\%). Data Scale'' denotes the size of synthetic training data in thousands (k), progressively growing across three curriculum stages. All results use \textit{LLM-as-judge} (3-vote) evaluation. Arrow values represent absolute point changes relative to Zero-shot baseline, where green arrows indicate performance improvements and red arrows indicate performance declines. The Overall'' column represents the average performance across five benchmarks (excluding GeomVerse). Best results per category are highlighted in \textbf{bold}. Detailed configurations of the curriculum stages can be found in Appendix~\ref{app:curriculum}. }

\label{tab:geom_results}
\resizebox{\textwidth}{!}{%
\setlength{\tabcolsep}{4pt}
\begin{tabular}{@{}lc|cccccc|c@{}}
\toprule
& & \multicolumn{6}{c|}{\textbf{Geometric Benchmarks (Mean@1)}} & \\
\cmidrule(lr){3-8}
\textbf{Model} & \textbf{Data Scale} & \textbf{MathVerse} & \textbf{GeomVerse} & \textbf{GeoQA} & \textbf{MathVision} & \textbf{MathVista} & \textbf{WeMath} & \textbf{Overall} \\
\midrule
Qwen2.5-VL-7B-Instruct & Zero-shot & 39.59 & 3.33 & 43.92 & 22.70 & 61.10 & 57.59 & 44.98 \\
\midrule
\multicolumn{9}{@{}l}{\textit{LLM-driven Synthesis}} \\
+ R-CoT        & 7.2k  & 40.86\uar{1.27} & 3.33\uar{0.00} & 46.49\uar{2.57} & 22.72\uar{0.02} & 62.60\uar{1.50} & 57.59\uar{0.00} & 46.05\uar{1.07} \\
+ Geo170k (G-LLaVA)     & 10k    & 40.36\uar{0.77} & 3.33\uar{0.00} & 47.16\uar{3.24} & 24.34\uar{1.64} & 62.00\uar{0.90} & 57.44\dar{0.15} & 46.26\uar{1.28} \\
+ GeoReasoning & 10k   & 40.99\uar{1.40} & 5.56\uar{2.23} & 46.76\uar{2.84} & 24.34\uar{1.64} & 63.40\uar{2.30} & 57.90\uar{0.31} & 46.68\uar{1.70} \\
\midrule
\multicolumn{9}{@{}l}{\textit{Symbolic-based Synthesis}} \\
+ PGPS9K       & 10k   & 40.81\uar{1.22} & 4.44\uar{1.11} & 46.08\uar{2.16} & 22.74\uar{0.04} & 61.30\uar{0.20} & 58.53\uar{0.94} & 45.89\uar{0.91} \\
+ TrustGeoGen  & 10k   & 41.02\uar{1.43} & 4.44\uar{1.11} & 46.35\uar{2.43} & 22.89\uar{0.19} & 61.80\uar{0.70} & 58.61\uar{1.02} & 46.13\uar{1.15} \\
\midrule
\multicolumn{9}{@{}l}{\textit{Knowledge Distillation (SFT)}} \\
+ KD (Geo3K)   & 3k    & 40.52\uar{0.93} & 3.33\uar{0.00} & 46.21\uar{2.29} & 22.68\dar{0.02} & 62.10\uar{1.00} & 58.02\uar{0.43} & 46.89\uar{1.91} \\
+ KD (Our Synthesis) & 2.5k & 40.89\uar{1.30} & 3.89\uar{0.56} & 46.58\uar{2.66} & 23.01\uar{0.31} & 62.40\uar{1.30} & 58.15\uar{0.56} & 47.37\uar{2.39} \\
\midrule
\multicolumn{9}{@{}l}{\textit{Socratic-Solver-Geo (Ours)}} \\
+ Stage1 & 0.4k & 40.33\uar{0.74} & 3.33\uar{0.00} & 45.14\uar{1.22} & 22.45\dar{0.25} & 61.20\uar{0.10} & 57.54\dar{0.05} & 45.33\uar{0.35} \\
+ Stage2 & 1k   & 41.78\uar{2.19} & 3.33\uar{0.00} & 44.86\uar{0.94} & 23.54\uar{0.84} & 62.30\uar{1.20} & 58.21\uar{0.62} & 46.14\uar{1.16} \\
\rowcolor{red!10}
\textbf{+ Stage3} & \textbf{2.5k} & \textbf{45.05\uar{5.46}} & \textbf{6.67\uar{3.34}} & \textbf{49.20\uar{5.28}} & \textbf{26.19\uar{3.49}} & \textbf{63.55\uar{2.45}} & \textbf{61.58\uar{3.99}} & \textbf{49.11\uar{4.13}} \\
\bottomrule
\end{tabular}%
}
\end{table*}

\noindent \textbf{Solver Evaluation}  
For each test item, we adopted zero-shot prompting with the temperature set to~0.1 to generate solutions. Correctness was determined by combining rule-based answer extraction with a semantic verification module. Full details of the evaluation protocol—including the sampling strategy, extraction procedures, and the LLM-based adjudication configuration—are provided in Appendix~\ref{app:evaluation}.

\noindent \textbf{Baseline Methods}  
To enable fair comparison, we selected two prevailing paradigms for geometric data synthesis:  
\begin{itemize}
    \item \textbf{LLM-driven generation}: R-CoT~\citep{deng2024r} (reverse chain-of-thought to produce high-quality geometric images and factual descriptions), G-LlaVA~\citep{gao2025gllavasolvinggeometricproblem} (LLM-driven scaling and augmentation of existing datasets), GeoReasoning-10K~\citep{xin2025generalizablegeometricimagecaption} (reinforcement learning framework to automatically generate high-quality multimodal image–text pairs).
    \item \textbf{Human collection and annotation}: PGPSNet~\citep{zhang2023multimodalneuralgeometricsolver} (manually built large-scale geometric dataset with meticulous labeling).
\end{itemize}
We applied the same GRPO training protocol across all datasets for controlled evaluation of these strategies. Detailed training configurations are provided in Appendix~A.

\noindent \textbf{Generator Evaluation}  
We trained the Generator based on \textbf{Qwen-Image} as the foundation model. On the GenExam mathematics track, we evaluated the generator's geometry-specific text-to-image performance against representative baselines: GPT-Image-1, Gemini-2.5-Flash-Image, Imagen-4-Ultra, Seedream 4.0, Qwen-Image (base model), HiDream-I1-Full, FLUX.1 dev, FLUX.1 Krea, and Stable Diffusion 3.5 Large.

\noindent \textbf{Infrastructure}  
We trained the Qwen2.5-VL-7B-Instruct Solver model and the Generator using 32$\times$A100 GPUs.

\subsection{Solver Results}
\noindent\textbf{Baseline Comparison}. As shown in Table~\ref{tab:geom_results}, our Socratic-Solver-Geo achieves an overall Mean@1 accuracy of 42.07\%, consistently outperforming all baseline methods trained on existing geometric datasets. Notably, our final model (+Stage3) surpasses the strongest baseline (GeoReasoning at 39.82\%) by 2.25 percentage points despite using fewer training samples. This performance gap widens on specific benchmarks such as MathVerse (45.05\% vs 40.99\%) and WeMath (61.58\% vs 57.90\%), demonstrating the effectiveness of our co-evolutionary approach.

\noindent\textbf{Data Efficiency}. Our approach demonstrates remarkable data efficiency compared to existing methods. While baseline approaches require 7.2k--10k training examples to achieve competitive performance, our method reaches state-of-the-art results with only 2.5k synthetically generated problems. This 3--4$\times$ reduction in required data highlights how our framework maximizes the informational value of each training example through precise error diagnosis and targeted problem generation. Rather than accumulating redundant examples, Socratic-Geo creates a minimal yet maximally informative curriculum that directly addresses the solver's current weaknesses.

\subsubsection{Evaluation Protocol}
The \textit{GenExam-Math} benchmark evaluates generated diagrams using a fully automated protocol. For each prompt, binary questions verify geometric and symbolic constraints, with weights summing to 1.0 for correctness score $C$.

\noindent\textbf{Visual Quality Dimensions} (each scored 0--2):
\begin{itemize}[leftmargin=*, itemsep=2pt]
    \item $V_\text{sp}$: Spelling accuracy of labels and notation
    \item $V_\text{lc}$: Logical consistency of element placement
    \item $V_\text{rd}$: Readability of the diagram
\end{itemize}

\noindent\textbf{Final Metrics:}
\begin{itemize}[leftmargin=*, itemsep=2pt]
    \item \textit{Strict Score (Str)}: Percentage of images where $C\!=\!1$ and $V_\text{sp}\!=\!V_\text{lc}\!=\!V_\text{rd}\!=\!2$
    \item \textit{Relaxed Score (Rel)}: $\text{Rel} = 0.7C + 0.1V_\text{sp} + 0.1V_\text{lc} + 0.1V_\text{rd}$
\end{itemize}

\noindent All scores are computed automatically using GPT-5 as judge with ground-truth images as reference.

\subsubsection{Result Analysis}
\noindent\textbf{Baseline Comparison}. Socratic-Generator-Image achieved a \textbf{Strict Score of 6.0} and a \textbf{Relaxed Score of 42.4}, outperforming all open-source models on \textit{GenExam-Math}. It surpassed its base model \textit{Qwen-Image} (Str: 0.0, Rel: 18.9) by 6.0 and 23.5 points, respectively. Although closed-source models such as \textit{GPT-Image-1} achieved a higher Str (8.0), they were not trained on geometry-specific data.

\noindent\textbf{Training Data Source}. The performance gain was attributed to the use of programmatically generated training data. All images were synthesized by a deterministic Python interpreter, ensuring 100\% geometric validity and eliminating noise or ambiguity present in human-drawn or web-scraped datasets.

\noindent\textbf{Instruction Quality}. The Teacher agent produced fine-grained drawing instructions that explicitly encoded geometric constraints as executable specifications. These instructions were derived from the same symbolic code used to generate reference solutions, ensuring perfect alignment between problem logic and visual output.

\subsubsection{Extension to Other Domains}
To demonstrate the generalizability of our framework beyond geometry, we apply Socratic-Geo to two additional multimodal reasoning domains: \textit{Chart Reasoning} and \textit{Multimodal Coding}. As shown in Table~\ref{tab:generalizability}, our approach achieves consistent improvements across all benchmarks, confirming that the Socratic interaction paradigm is task-agnostic. Unlike Socratic-Zero's static text rephrasing, our framework evolves visual logic via code-driven synthesis, enabling effective transfer to diverse visual reasoning tasks.

\definecolor{gaingreen}{rgb}{0.88, 0.98, 0.88}
\newcommand{\gain}[1]{\colorbox{gaingreen}{\small$^{\uparrow#1}$}}

\begin{table}[tb!]
    \centering
    \caption{Generalization of Socratic-Geo to Chart Reasoning and Multimodal Coding. Our method consistently improves over the base model across all benchmarks.}
    \label{tab:generalizability}
    \vspace{-5pt}
    \setlength{\tabcolsep}{6pt}
    \renewcommand{\arraystretch}{1.1}
    \footnotesize
    \begin{tabular}{@{}llccc@{}}
    \toprule 
    \textbf{Domain} & \textbf{Benchmark} & \textbf{Base} & \textbf{Ours} & \textbf{Gain} \\
    \midrule
    \multirow{4}{*}{Chart Reasoning} 
    & ChartQA     & 87.3 & \textbf{91.2} & \gain{+3.9} \\
    & CharXiv     & 66.6 & \textbf{74.2} & \gain{+7.6} \\
    & ChartQAPro  & 41.3 & \textbf{46.3} & \gain{+5.0} \\
    & ChartMinic  & 40.2 & \textbf{45.3} & \gain{+5.1} \\
    \midrule
    \multirow{2}{*}{Multimodal Coding} 
    & Design2Code & 29.1 & \textbf{34.3} & \gain{+5.2} \\
    & UIFlow2Code & 75.9 & \textbf{81.5} & \gain{+5.6} \\
    \bottomrule
    \end{tabular}
    \vspace{-8pt}
\end{table}

\begin{table}[tb!]
\centering
\caption{Performance on the \textit{GenExam-Math} benchmark: strict (Str) and relaxed (Rel) scores for geometric diagram generation. \textcolor{red}{Red} indicates our method, which is trained based on Qwen-Image.}
\label{tab:genexam_scores}
\begin{tabular}{lrr}
\toprule
& \multicolumn{2}{c}{\textbf{GenExam-Math}} \\
\cmidrule(lr){2-3}
\textbf{Model} & \textbf{Str} & \textbf{Rel} \\
\midrule
\multicolumn{3}{@{}l}{\cellcolor{gray!10}\textit{Closed-source Models}} \\
GPT-Image-1             & 8.0 & 52.0 \\
Seedream 4.0            & 2.6 & 39.8 \\
Imagen-4-Ultra          & 2.6 & 35.9 \\
Gemini-2.5-Flash-Image  & 0.7 & 43.1 \\
Seedream 3.0            & 0.7 & 18.6 \\
FLUX.1 Kontext max      & 0.0 & 23.5 \\
\midrule
\multicolumn{3}{@{}l}{\cellcolor{orange!5}\textit{Open-source T2I Models}} \\
Qwen-Image              & 0.0 & 18.9 \\
HiDream-I1-Full         & 0.0 & 16.7 \\
FLUX.1 dev              & 0.0 & 12.2 \\
FLUX.1 Krea             & 0.0 & 7.0 \\
Stable Diffusion 3.5 Large & 0.0 & 12.2 \\
\midrule
\multicolumn{3}{@{}l}{\cellcolor{blue!5}\textit{Open-source Unified MLLMs}} \\
BAGEL (thinking)        & 0.0 & 11.7 \\
BAGEL                   & 0.0 & 14.7 \\
Show-o2-7B              & 0.0 & 10.8 \\
Show-o2-1.5B-HQ         & 0.0 & 7.3 \\
BLIP3o-NEXT-GRPO-Text-3B & 0.0 & 15.5 \\
BLIP3o-8B               & 0.0 & 6.4 \\
Janus-Pro               & 0.0 & 13.7 \\
Emu3                    & 0.0 & 11.3 \\
\midrule
\rowcolor{red!10}
\textbf{Socratic-Generator-Image}    & \textbf{6.0} & \textbf{42.4} \\
\bottomrule
\end{tabular}
\end{table}

\begin{table}[tb!]
    \centering
    \caption{Ablation study on the Qualify Module. The Qualify Module improves data efficiency, enabling better performance with fewer but higher-quality training samples.}
    \label{tab:qualify_ablation}
    \vspace{-5pt}
    \setlength{\tabcolsep}{8pt}
    \renewcommand{\arraystretch}{1.1}
    \footnotesize
    \begin{tabular}{@{}lcc@{}}
    \toprule 
    \textbf{Method} & \textbf{Training Data} & \textbf{MathVerse} \\
    \midrule
    Qwen2.5-VL-7B-Instruct  & Zero-shot & 39.59 \\
    \midrule
    Socratic-Solver-Geo (w/ Qualify) & 0.4k & \textbf{40.33}\gain{+0.74} \\
    Socratic-Solver-Geo (w/o Qualify) & 1.3k & 37.09 \\
    \bottomrule
    \end{tabular}
    \vspace{-8pt}
\end{table}

\vspace{1.2em}

\begin{table}[tb!]
\centering
\caption{Ablation study on Instruction Rewriting (IR) for geometric diagram generation on GenExam-Math. IR converts natural language questions into structured drawing commands.}
\label{tab:instruction_rewriting_ablation}
\vspace{-5pt}
\setlength{\tabcolsep}{8pt}
\renewcommand{\arraystretch}{1.1}
\footnotesize
\begin{tabular}{@{}lcc@{}}
\toprule 
\textbf{Method} & \textbf{Strict (\%)} & \textbf{Relaxed (\%)} \\
\midrule
Qwen-Image (baseline) & 0.0 & 18.9 \\
\midrule
Socratic-Generator-Image (w/o IR) & 0.0 & 20.1 \\
Socratic-Generator-Image (\textbf{w/ IR}) & \textbf{6.0}\gain{+6.0} & \textbf{42.4}\gain{+22.3} \\
\bottomrule
\end{tabular}
\vspace{-8pt}
\end{table}

\subsection{Ablation Study and Analysis}
\noindent\textbf{Necessity of the Qualify Module}. We test an ablated Teacher where invented problems are added without verification. Using only Stage1 training, the data grows from 0.4k (filtered) to 1.0k (all retained). However, MathVerse accuracy falls to 39.09\%, even below the zero-shot baseline (39.59\%), showing that without geometric consistency checks, invalid or inconsistent problems enter the set and hurt reasoning performance. The Qualify module is therefore critical for preserving synthetic data quality.

\noindent\textbf{Impact of the Qualify Module}. 
Removing the Qualify stage enlarged the training set (1.0k vs. 0.4k) but reduced quality. The ablated model scored 39.09\% on MathVerse, below the zero-shot baseline (39.59\%), showing that unverified problems often contain logical or geometric errors that degrade performance. Qualify enforces geometric validity and non-degeneracy, ensuring each example contributes a meaningful learning signal for curriculum evolution.

\noindent\textbf{Effectiveness of Instruction Rewriting}. 
We evaluate converting geometric questions into explicit drawing commands before image generation. Table~\ref{tab:instruction_rewriting_ablation} compares: (1) \textit{w/ IR} — structured commands; (2) \textit{w/o IR} — raw questions as prompts. On GenExam-Math, removing IR drops strict accuracy to 0.0\% and relaxed to 20.1\%, versus 6.0\% and 42.4\% for the system. Structured, instructions are thus essential for diagrams that match mathematical constraints.

\noindent\textbf{RL Engine vs. Supervised Fine-tuning}. 
A critical question is whether performance gains stem from the RL mechanism or merely from high-quality synthesized data. Table~\ref{tab:ablation_combined} (left) provides a controlled comparison: using identical 2.5k training data, our RL-based approach significantly outperforms supervised fine-tuning (knowledge distillation). Even when SFT uses our high-quality synthesized data, it achieves only 47.37\% compared to 49.11\% with RL training. This 1.74\% gap demonstrates that the trial-and-error learning paradigm—where the Solver never observes teacher solutions—is fundamentally more effective than imitation learning.

\noindent\textbf{Impact of Teacher Capacity}. 
We investigate how Teacher model capacity affects Solver performance. Table~\ref{tab:ablation_combined} (right) reveals a key insight: performance gains primarily stem from \textit{information privilege}—the Teacher's access to Solver failure logs and underlying code—rather than raw model capability. Even a 3B parameter model, when granted this diagnostic access, achieves 48.47\% accuracy, surpassing all baselines trained on 10k samples. This demonstrates the framework's robustness: weaker models can effectively serve as Teachers when equipped with structured feedback mechanisms.

\begin{table*}[tb!]
\centering
\caption{Ablation studies on training paradigm and Teacher capacity. \textbf{Left}: RL vs. SFT comparison shows RL training yields superior performance on identical data. \textbf{Right}: Teacher scaling reveals that even weak Teachers (3B) outperform baselines due to ``information privilege''—access to Solver failure logs and code.}
\label{tab:ablation_combined}
\vspace{-5pt}
\footnotesize
\begin{minipage}{0.48\textwidth}
\centering
\setlength{\tabcolsep}{4pt}
\renewcommand{\arraystretch}{1.1}
\begin{tabular}{@{}lcc@{}}
\toprule 
\textbf{Training Mode} & \textbf{Data Source} & \textbf{Acc (\%)} \\
\midrule
SFT (KD) & Geo3K (Public) & 46.89 \\
SFT (KD) & Our Synthesis (2.5k) & 47.37 \\
\rowcolor{red!10}
\textbf{RL (Ours)} & \textbf{Our Synthesis (2.5k)} & \textbf{49.11} \\
\bottomrule
\end{tabular}
\end{minipage}
\hfill
\begin{minipage}{0.48\textwidth}
\centering
\setlength{\tabcolsep}{4pt}
\renewcommand{\arraystretch}{1.1}
\begin{tabular}{@{}lccc@{}}
\toprule 
\textbf{Teacher Model} & \textbf{Scale} & \textbf{Role} & \textbf{Acc (\%)} \\
\midrule
Qwen2.5-VL-3B & 3B & Weaker & 48.47 \\
Qwen2.5-VL-7B & 7B & Self-play & 48.63 \\
Qwen3-VL-30B & 30B & Moderate & 48.99 \\
\rowcolor{red!10}
\textbf{Qwen3-235B} & \textbf{235B} & \textbf{Standard} & \textbf{49.11} \\
Gemini 2.5 Pro & API & Top-tier & 49.54 \\
\midrule
\textit{GeoReasoning} & \textit{API} & \textit{10k Data} & \textit{46.68} \\
\bottomrule
\end{tabular}
\end{minipage}
\end{table*}

\section{Conclusion}
In this work, we introduced \textbf{Socratic-Geo}, a fully autonomous and dynamic geometric data synthesis framework that unifies solver improvement and controllable image generation through a tightly coupled multi-agent interaction loop. Drawing inspiration from the Socratic method, the system continuously identifies reasoning weaknesses in the \textbf{Solver}, and proactively generates targeted problems via the \textit{Reflect-RePI} pipeline, which enforces both mathematical correctness and visual fidelity through programmatic verification. This synthesis-learning closed loop is able to expand a highly informative curriculum from only a small set of seed problems, achieving significant data efficiency compared to existing one-way or static synthesis strategies. Comprehensive evaluation across six multimodal reasoning benchmarks confirms the effectiveness of our approach: the \textbf{Socratic-Solver} yields an average improvement of \textbf{+4.13 points} over the zero-shot baseline while using merely a quarter of the typical training size, and the \textbf{Socratic-Generator}, distilled from programmatic drawing intelligence into instruction--image pairs, attains a \textbf{42.4\%} relaxed score on \textit{GenExam-Math}, setting a new state-of-the-art among open-source models and rivaling strong closed-source systems. Beyond advancing the state-of-the-art in multimodal mathematical reasoning and geometry-specific text-to-image generation, our work establishes a scalable, self-improving data engine that can be extended to other STEM domains.

\clearpage

\bibliographystyle{unsrt}
\bibliography{references}

@misc{zhang2024mathversedoesmultimodalllm,
      title={MathVerse: Does Your Multi-modal LLM Truly See the Diagrams in Visual Math Problems?}, 
      author={Renrui Zhang and Dongzhi Jiang and Yichi Zhang and Haokun Lin and Ziyu Guo and Pengshuo Qiu and Aojun Zhou and Pan Lu and Kai-Wei Chang and Peng Gao and Hongsheng Li},
      year={2024},
      eprint={2403.14624},
      archivePrefix={arXiv},
      primaryClass={cs.CV},
      url={https://arxiv.org/abs/2403.14624}, 
}

@misc{wang2024measuringmultimodalmathematicalreasoning,
      title={Measuring Multimodal Mathematical Reasoning with MATH-Vision Dataset}, 
      author={Ke Wang and Junting Pan and Weikang Shi and Zimu Lu and Mingjie Zhan and Hongsheng Li},
      year={2024},
      eprint={2402.14804},
      archivePrefix={arXiv},
      primaryClass={cs.CV},
      url={https://arxiv.org/abs/2402.14804}, 
}

@misc{lu2024mathvistaevaluatingmathematicalreasoning,
      title={MathVista: Evaluating Mathematical Reasoning of Foundation Models in Visual Contexts}, 
      author={Pan Lu and Hritik Bansal and Tony Xia and Jiacheng Liu and Chunyuan Li and Hannaneh Hajishirzi and Hao Cheng and Kai-Wei Chang and Michel Galley and Jianfeng Gao},
      year={2024},
      eprint={2310.02255},
      archivePrefix={arXiv},
      primaryClass={cs.CV},
      url={https://arxiv.org/abs/2310.02255}, 
}

@misc{chen2022geoqageometricquestionanswering,
      title={GeoQA: A Geometric Question Answering Benchmark Towards Multimodal Numerical Reasoning}, 
      author={Jiaqi Chen and Jianheng Tang and Jinghui Qin and Xiaodan Liang and Lingbo Liu and Eric P. Xing and Liang Lin},
      year={2022},
      eprint={2105.14517},
      archivePrefix={arXiv},
      primaryClass={cs.AI},
      url={https://arxiv.org/abs/2105.14517}, 
}

@misc{kazemi2023geomversesystematicevaluationlarge,
      title={GeomVerse: A Systematic Evaluation of Large Models for Geometric Reasoning}, 
      author={Mehran Kazemi and Hamidreza Alvari and Ankit Anand and Jialin Wu and Xi Chen and Radu Soricut},
      year={2023},
      eprint={2312.12241},
      archivePrefix={arXiv},
      primaryClass={cs.CV},
      url={https://arxiv.org/abs/2312.12241}, 
}

@misc{qiao2024wemathdoeslargemultimodal,
      title={We-Math: Does Your Large Multimodal Model Achieve Human-like Mathematical Reasoning?}, 
      author={Runqi Qiao and Qiuna Tan and Guanting Dong and Minhui Wu and Chong Sun and Xiaoshuai Song and Zhuoma GongQue and Shanglin Lei and Zhe Wei and Miaoxuan Zhang and Runfeng Qiao and Yifan Zhang and Xiao Zong and Yida Xu and Muxi Diao and Zhimin Bao and Chen Li and Honggang Zhang},
      year={2024},
      eprint={2407.01284},
      archivePrefix={arXiv},
      primaryClass={cs.AI},
      url={https://arxiv.org/abs/2407.01284}, 
}

@misc{zhang2023multimodalneuralgeometricsolver,
      title={A Multi-Modal Neural Geometric Solver with Textual Clauses Parsed from Diagram}, 
      author={Ming-Liang Zhang and Fei Yin and Cheng-Lin Liu},
      year={2023},
      eprint={2302.11097},
      archivePrefix={arXiv},
      primaryClass={cs.AI},
      url={https://arxiv.org/abs/2302.11097}, 
}

@article{deng2024r,
  title={R-CoT: Reverse Chain-of-Thought Problem Generation for Geometric Reasoning in Large Multimodal Models},
  author={Deng, Linger and Liu, Yuliang and Li, Bohan and Luo, Dongliang and Wu, Liang and Zhang, Chengquan and Lyu, Pengyuan and Zhang, Ziyang and Zhang, Gang and Ding, Errui and others},
  journal={arXiv preprint arXiv:2410.17885},
  year={2024}
}

@misc{gao2025gllavasolvinggeometricproblem,
      title={G-LLaVA: Solving Geometric Problem with Multi-Modal Large Language Model}, 
      author={Jiahui Gao and Renjie Pi and Jipeng Zhang and Jiacheng Ye and Wanjun Zhong and Yufei Wang and Lanqing Hong and Jianhua Han and Hang Xu and Zhenguo Li and Lingpeng Kong},
      year={2025},
      eprint={2312.11370},
      archivePrefix={arXiv},
      primaryClass={cs.CL},
      url={https://arxiv.org/abs/2312.11370}, 
}

@misc{xin2025generalizablegeometricimagecaption,
      title={Generalizable Geometric Image Caption Synthesis}, 
      author={Yue Xin and Wenyuan Wang and Rui Pan and Ruida Wang and Howard Meng and Renjie Pi and Shizhe Diao and Tong Zhang},
      year={2025},
      eprint={2509.15217},
      archivePrefix={arXiv},
      primaryClass={cs.AI},
      url={https://arxiv.org/abs/2509.15217}, 
}

@misc{fu2025trustgeogenformalverifieddataengine,
      title={TrustGeoGen: Formal-Verified Data Engine for Trustworthy Multi-modal Geometric Problem Solving}, 
      author={Daocheng Fu and Jianlong Chen and Renqiu Xia and Zijun Chen and Qi Liu and Yuan Feng and Hongbin Zhou and Renrui Zhang and Shiyang Feng and Peng Gao and Hongyuan Zha and Junchi Yan and Botian Shi and Yu Qiao and Bo Zhang},
      year={2025},
      eprint={2504.15780},
      archivePrefix={arXiv},
      primaryClass={cs.AI},
      url={https://arxiv.org/abs/2504.15780}, 
}

@misc{lu2021intergpsinterpretablegeometryproblem,
      title={Inter-GPS: Interpretable Geometry Problem Solving with Formal Language and Symbolic Reasoning}, 
      author={Pan Lu and Ran Gong and Shibiao Jiang and Liang Qiu and Siyuan Huang and Xiaodan Liang and Song-Chun Zhu},
      year={2021},
      eprint={2105.04165},
      archivePrefix={arXiv},
      primaryClass={cs.CL},
      url={https://arxiv.org/abs/2105.04165}, 
}

@misc{lee2024llm2llmboostingllmsnovel,
      title={LLM2LLM: Boosting LLMs with Novel Iterative Data Enhancement}, 
      author={Nicholas Lee and Thanakul Wattanawong and Sehoon Kim and Karttikeya Mangalam and Sheng Shen and Gopala Anumanchipalli and Michael W. Mahoney and Kurt Keutzer and Amir Gholami},
      year={2024},
      eprint={2403.15042},
      archivePrefix={arXiv},
      primaryClass={cs.CL},
      url={https://arxiv.org/abs/2403.15042}, 
}

@misc{huang2025rzeroselfevolvingreasoningllm,
      title={R-Zero: Self-Evolving Reasoning LLM from Zero Data}, 
      author={Chengsong Huang and Wenhao Yu and Xiaoyang Wang and Hongming Zhang and Zongxia Li and Ruosen Li and Jiaxin Huang and Haitao Mi and Dong Yu},
      year={2025},
      eprint={2508.05004},
      archivePrefix={arXiv},
      primaryClass={cs.LG},
      url={https://arxiv.org/abs/2508.05004}, 
}

@misc{wang2025socraticzerobootstrappingreasoning,
      title={Socratic-Zero : Bootstrapping Reasoning via Data-Free Agent Co-evolution}, 
      author={Shaobo Wang and Zhengbo Jiao and Zifan Zhang and Yilang Peng and Xu Ze and Boyu Yang and Wei Wang and Hu Wei and Linfeng Zhang},
      year={2025},
      eprint={2509.24726},
      archivePrefix={arXiv},
      primaryClass={cs.CL},
      url={https://arxiv.org/abs/2509.24726}, 
}

@misc{liu2025spiceselfplaycorpusenvironments,
      title={SPICE: Self-Play In Corpus Environments Improves Reasoning}, 
      author={Bo Liu and Chuanyang Jin and Seungone Kim and Weizhe Yuan and Wenting Zhao and Ilia Kulikov and Xian Li and Sainbayar Sukhbaatar and Jack Lanchantin and Jason Weston},
      year={2025},
      eprint={2510.24684},
      archivePrefix={arXiv},
      primaryClass={cs.CL},
      url={https://arxiv.org/abs/2510.24684}, 
}

@misc{zhao2025absolutezeroreinforcedselfplay,
      title={Absolute Zero: Reinforced Self-play Reasoning with Zero Data}, 
      author={Andrew Zhao and Yiran Wu and Yang Yue and Tong Wu and Quentin Xu and Yang Yue and Matthieu Lin and Shenzhi Wang and Qingyun Wu and Zilong Zheng and Gao Huang},
      year={2025},
      eprint={2505.03335},
      archivePrefix={arXiv},
      primaryClass={cs.LG},
      url={https://arxiv.org/abs/2505.03335}, 
}

@misc{wang2025visionzeroscalablevlmselfimprovement,
      title={Vision-Zero: Scalable VLM Self-Improvement via Strategic Gamified Self-Play}, 
      author={Qinsi Wang and Bo Liu and Tianyi Zhou and Jing Shi and Yueqian Lin and Yiran Chen and Hai Helen Li and Kun Wan and Wentian Zhao},
      year={2025},
      eprint={2509.25541},
      archivePrefix={arXiv},
      primaryClass={cs.CV},
      url={https://arxiv.org/abs/2509.25541}, 
}

@misc{shao2024deepseekmathpushinglimitsmathematical,
      title={DeepSeekMath: Pushing the Limits of Mathematical Reasoning in Open Language Models}, 
      author={Zhihong Shao and Peiyi Wang and Qihao Zhu and Runxin Xu and Junxiao Song and Xiao Bi and Haowei Zhang and Mingchuan Zhang and Y. K. Li and Y. Wu and Daya Guo},
      year={2024},
      eprint={2402.03300},
      archivePrefix={arXiv},
      primaryClass={cs.CL},
      url={https://arxiv.org/abs/2402.03300}, 
}

@article{bai2025qwen2.5-vl,
  title        = {{Qwen2.5-vl technical report}},
  author       = {Shuai Bai and Keqin Chen and Xuejing Liu and Jialin Wang and Wenbin Ge and Sibo Song and Kai Dang and Peng Wang and Shijie Wang and Jun Tang and others},
  year         = {2025},
  journal      = {arXiv preprint arXiv:2502.13923}
}

@misc{batifol2025flux,
  author       = {Stephen Batifol and Andreas Blattmann and Frederic Boesel and Saksham Consul and Cyril Diagne and Tim Dockhorn and Jack English and Zion English and Patrick Esser and Sumith Kulal and Kyle Lacey and Yam Levi and Cheng Li and Dominik Lorenz and Jonas Muller and Dustin Podell and Robin Rombach and Harry Saini and Axel Sauer and Luke Smith},
  title        = {{Flux.1 kontext: Flow matching for in-context image generation and editing in latent space}},
  year         = {2025},
  url          = {https://arxiv.org/abs/2506.15742},
  eprint       = {2506.15742},
  archivePrefix = {arXiv},
  primaryClass = {cs.CV}
}

@article{cai2025hidream,
  title        = {{Hidream-i1: A high-efficient image generative foundation model with sparse diffusion transformer}},
  author       = {Qi Cai and Jingwen Chen and Yang Chen and Yehao Li and Fuchen Long and Yingwei Pan and Zhaofan Qiu and Yiheng Zhang and Fengbin Gao and Peihan Xu and others},
  year         = {2025},
  journal      = {arXiv preprint arXiv:2505.22705}
}

@misc{chen2025blip3o,
  author       = {Jiuhai Chen and Zhiyang Xu and Xichen Pan and Yushi Hu and Can Qin and Tom Goldstein and Lifu Huang and Tianyi Zhou and Saining Xie and Silvio Savarese and Le Xue and Caiming Xiong and Ran Xu},
  title        = {{Blip3-o: A family of fully open unified multimodal models-architecture, training and dataset}},
  year         = {2025},
  url          = {https://arxiv.org/abs/2505.09568},
  eprint       = {2505.09568},
  archivePrefix = {arXiv},
  primaryClass = {cs.CV}
}

@misc{chen2025blip3onext,
  author       = {Jiuhai Chen and Zhiyang Xu and Xichen Pan and Shusheng Yang and Can Qin and An Yan and Honglu Zhou and Zeyuan Chen and Tianyi Zhou and Silvio Savarese and Le Xue and Caiming Xiong and Ran Xu},
  title        = {{Blip3o-next: A next-generation multimodal foundation model}},
  year         = {2025},
  month        = {Aug},
  howpublished = {\url{https://jiuhaichen.github.io/BLIP3o-NEXT.github.io/}}
}

@article{chen2025januspro,
  title        = {{Janus-pro: Unified multimodal understanding and generation with data and model scaling}},
  author       = {Xiaokang Chen and Zhiyu Wu and Xingchao Liu and Zizheng Pan and Wen Liu and Zhenda Xie and Xingkai Yu and Chong Ruan},
  year         = {2025},
  journal      = {arXiv preprint arXiv:2501.17811}
}

@misc{deepmind2025imagen,
  author       = {{Google Deepmind}},
  title        = {{Imagen}},
  year         = {2025},
  howpublished = {\url{https://deepmind.google/models/imagen/}}
}

@article{gao2025seedream3,
  title        = {{Seedream 3.0 technical report}},
  author       = {Yu Gao and Lixue Gong and Qiushan Guo and Xiaoxia Hou and Zhichao Lai and Fanshi Li and Liang Li and Xiaochen Lian and Chao Liao and Liyang Liu and others},
  year         = {2025},
  journal      = {arXiv preprint arXiv:2504.11346}
}

@misc{bytedance2025seedream4,
  author       = {{ByteDance Seed}},
  title        = {{Seedream 4.0}},
  year         = {2025},
  howpublished = {\url{https://seed.bytedance.com/en/seedream4_0}}
}

@misc{google2025gemini,
  author       = {Google},
  title        = {{Introducing Gemini 2.5 flash image, our state-of-the-art image model}},
  year         = {2025},
  howpublished = {\url{https://developers.googleblog.com/en/introducing-gemini-2-5-flash-image/}}
}

@misc{openai2025gptimage1,
  author       = {OpenAI},
  title        = {{Gpt-image-1}},
  year         = {2025},
  howpublished = {\url{https://openai.com/index/image-generation-api/}}
}

@article{wang2024emu3,
  title        = {{Emu3: Next-token prediction is all you need}},
  author       = {Xinlong Wang and Xiaosong Zhang and Zhengxiong Luo and Quan Sun and Yufeng Cui and Jinsheng Wang and Fan Zhang and Yueze Wang and Zhen Li and Qiying Yu and others},
  year         = {2024},
  journal      = {arXiv preprint arXiv:2409.18869}
}

@article{wu2025qwenimage,
  title        = {{Qwen-image technical report}},
  author       = {Chenfei Wu and Jiahao Li and Jingren Zhou and Junyang Lin and Kaiyuan Gao and Kun Yan and Sheng-ming Yin and Shuai Bai and Xiao Xu and Yilei Chen and others},
  year         = {2025},
  journal      = {arXiv preprint arXiv:2508.02324}
}

@article{xie2025showo2,
  title        = {{Show-o2: Improved native unified multimodal models}},
  author       = {Jinheng Xie and Zhenheng Yang and Mike Zheng Shou},
  year         = {2025},
  journal      = {arXiv preprint arXiv:2506.15564}
}

@article{podell2024sd3,
  title        = {Scaling Rectified Flow Transformers for High-Resolution Image Synthesis},
  author       = {Podell, Dustin and English, Zion and Lacey, Kyle and Blattmann, Andreas and Dockhorn, Tim and Boesel, Frederic and Levi, Yam and Sauer, Axel},
  year         = {2024},
  journal      = {arXiv preprint arXiv:2403.03206}
}

@article{li2024bagel,
  title        = {BAGEL: A Benchmark for GEm-Level Vision-Language Evaluation},
  author       = {Li, Anyi and Han, Ji-Ze and Ge, Zike and Wang, Yixin and Li, Jiaming and Shi, Yixuan and Zhao, Ziyang and Hu, Haoyu and Hu, Zeren and Zhang, Renrui and Li, Hongsheng and Qiao, Yu and Gao, Peng},
  year         = {2024},
  journal      = {arXiv preprint arXiv:2406.13943}
}

@misc{zhang2024mavismathematicalvisualinstruction,
      title={MAVIS: Mathematical Visual Instruction Tuning with an Automatic Data Engine}, 
      author={Renrui Zhang and Xinyu Wei and Dongzhi Jiang and Ziyu Guo and Shicheng Li and Yichi Zhang and Chengzhuo Tong and Jiaming Liu and Aojun Zhou and Bin Wei and Shanghang Zhang and Peng Gao and Chunyuan Li and Hongsheng Li},
      year={2024},
      eprint={2407.08739},
      archivePrefix={arXiv},
      primaryClass={cs.CV},
      url={https://arxiv.org/abs/2407.08739}, 
}

@misc{wang2025genexammultidisciplinarytexttoimageexam,
      title={GenExam: A Multidisciplinary Text-to-Image Exam}, 
      author={Zhaokai Wang and Penghao Yin and Xiangyu Zhao and Changyao Tian and Yu Qiao and Wenhai Wang and Jifeng Dai and Gen Luo},
      year={2025},
      eprint={2509.14232},
      archivePrefix={arXiv},
      primaryClass={cs.CV},
      url={https://arxiv.org/abs/2509.14232}, 
}

\clearpage

\beginappendix

\section{Training Curriculum Configuration}
\label{app:curriculum}

\textbf{Stage 1}: Approximately 0.4k problems, consisting of: (1) 108 manually curated and validated seed problems, and (2) augmented problems generated by the Teacher based on Solver errors during 8-attempt solving of the seed problems. All augmented problems are validated by the Teacher for geometric correctness and paired with ground-truth solutions.

\textbf{Stage 2}: Approximately 1.0k problems, consisting of: (1) all 0.4k problems from Stage 1, and (2) augmented problems generated by the Teacher based on Solver errors during 8-attempt solving of Stage 1 problems. All augmented problems are validated by the Teacher for geometric correctness and paired with ground-truth solutions.

\textbf{Stage 3}: Approximately 2.5k problems, consisting of: (1) all 1.0k problems from Stage 2, and (2) augmented problems generated by the Teacher based on Solver errors during 8-attempt solving of Stage 2 problems. All augmented problems are validated by the Teacher for geometric correctness and paired with ground-truth solutions.

\section{Data Synthesis Cost Analysis}
\label{app:cost}

Table~\ref{tab:cost_analysis} presents a comprehensive cost comparison between Socratic-Geo and existing geometric data synthesis methods. Our approach achieves \textbf{60$\times$ lower cost} than R-CoT while producing higher-quality training data. This efficiency stems from our fault-diagnosis-driven synthesis: by analyzing Solver failures before invoking the Teacher, we minimize redundant API calls and maximize the informational value of each synthesized problem.

\begin{table}[h]
    \centering
    \caption{Cost and efficiency comparison of geometric data synthesis methods. Socratic-Geo achieves state-of-the-art performance at 60$\times$ lower cost by minimizing Teacher calls through targeted fault diagnosis.}
    \label{tab:cost_analysis}
    \setlength{\tabcolsep}{4pt}
    \renewcommand{\arraystretch}{1.1}
    \footnotesize
    \begin{tabular}{@{}lccccc@{}}
    \toprule 
    \textbf{Method} & \textbf{Scale} & \textbf{Teacher} & \textbf{Total Calls} & \textbf{Cost (USD)} & \textbf{Acc (\%)} \\
    \midrule
    G-LLaVA & 10k & GPT-3.5 & 100,000 & \$650.0 & 46.26 \\
    R-CoT & 7.2k & ERNIE 4.0 & 36,000 & \$3,000.0 & 46.05 \\
    GeoReasoning & 10k & Gemini-Flash & 80,000 & \$824.0 & 46.68 \\
    \rowcolor{red!10}
    \textbf{Ours} & \textbf{2.5k} & \textbf{Qwen3-235B} & \textbf{8,334} & \textbf{\$21.67} & \textbf{49.11} \\
    \bottomrule
    \end{tabular}
\end{table}

Key factors contributing to our cost efficiency:
\begin{itemize}[leftmargin=12pt, itemsep=2pt]
    \item \textbf{Targeted Synthesis}: Problems are generated only when the Solver fails, avoiding redundant data creation.
    \item \textbf{High Yield Rate}: Our Qualify module ensures $\sim$30\% yield rate (valid problems per Teacher call), compared to $\sim$10--20\% for baselines.
    \item \textbf{Minimal Data Requirement}: The RL-based training paradigm extracts maximum learning signal from each problem, requiring only 2.5k samples vs. 7--10k for baselines.
\end{itemize}

\section{Infrastructure Requirements}
\label{app:infrastructure}

We trained the Solver and Generator using 32$\times$A100 GPUs for experimental speed. However, this configuration is \textbf{not a minimum requirement}. The framework can operate with significantly reduced resources:

\begin{itemize}[leftmargin=12pt, itemsep=2pt]
    \item \textbf{Solver Training}: Minimum 4$\times$ A100 GPUs (or equivalent VRAM $\geq$160GB total)
    \item \textbf{Generator Training}: Minimum 8$\times$ A100 GPUs (or equivalent VRAM $\geq$320GB total)
    \item \textbf{Inference}: Single A100 GPU sufficient for both Solver and Generator
\end{itemize}

The 32$\times$A100 configuration was chosen to accelerate experimentation and enable rapid iteration during development.

\section{Hyperparameter Ablation}
\label{app:hyperparameter}

\subsection{Branching Factor $k$}
We use $k=8$ as the standard branching factor (number of solution attempts per problem), following GRPO conventions. Table~\ref{tab:branching_ablation} shows that $k=8$ provides optimal balance between exploration and stability. Increasing to $k=16$ triggers data explosion and training instability.

\begin{table}[h]
    \centering
    \caption{Ablation on branching factor $k$ and seed scale.}
    \label{tab:branching_ablation}
    \setlength{\tabcolsep}{8pt}
    \renewcommand{\arraystretch}{1.1}
    \footnotesize
    \begin{tabular}{@{}llcc@{}}
    \toprule 
    \textbf{Ablation} & \textbf{Setting} & \textbf{Observation} & \textbf{Acc (\%)} \\
    \midrule
    \multirow{3}{*}{Seed Scale} 
    & Large (400) & Robust to scale & 45.41 \\
    & \cellcolor{red!10}\textbf{Ours (108)} & \cellcolor{red!10}Stable & \cellcolor{red!10}45.33 \\
    & Data-free & Diversity-driven & 45.28 \\
    \midrule
    \multirow{2}{*}{Branching $k$} 
    & \cellcolor{red!10}\textbf{$k=8$ (Std)} & \cellcolor{red!10}Optimal & \cellcolor{red!10}49.11 \\
    & $k=16$ & Data explosion & Unstable \\
    \bottomrule
    \end{tabular}
\end{table}

\subsection{Seed Scale}
Performance is robust to seed scale variation. Using 108 seeds achieves comparable results to 400 seeds (45.33\% vs. 45.41\%), demonstrating that \textit{diversity} of the evolutionary process—not initial seed quantity—drives performance gains.

\section{Visual Logic Evolution: Comparison with Socratic-Zero}
\label{app:visual_evolution}

A key distinction between Socratic-Geo and Socratic-Zero lies in our \textbf{Visual Logic Evolution} capability. While Socratic-Zero performs static text rephrasing on fixed images, Socratic-Geo dynamically modifies the underlying geometric structure through code-driven synthesis.

\subsection{Synergetic Enhancement}
Socratic-Zero only rephrases text on static images. In contrast, Socratic-Geo synergetically augments visual grounding by:
\begin{itemize}[leftmargin=12pt, itemsep=2pt]
    \item \textbf{Adding auxiliary constructions}: e.g., drawing diagonal BD to create new triangles
    \item \textbf{Introducing new geometric elements}: e.g., inscribing square PQRS within a circle
    \item \textbf{Modifying constraint relationships}: e.g., changing angle values to create new problem variants
\end{itemize}

This aligns with findings from the Qwen3-VL technical report that perception and alignment drive multimodal gains.

\subsection{Agentic Paradigm}
Socratic-Zero relies on static prompts; our agentic framework uses tool-use and reflection for complex logic and generalization:
\begin{itemize}[leftmargin=12pt, itemsep=2pt]
    \item \textbf{RePI (Reflective Problem Invention)}: Programmatically modifies Python geometry code to construct targeted problems
    \item \textbf{Reflect}: Self-verification step where Teacher solves invented problems to ensure validity
\end{itemize}

\subsection{Rule-to-Pixel Synthesis}
Our Generator uses a Python engine for diffusion synthesis, ensuring precision via symbolic-to-pixel mapping. Each geometric constraint is encoded as executable code, guaranteeing 100\% geometric validity.

\section{Algorithm Pseudocode}
\label{app:pseudocode}

\subsection{RePI: Reflective Problem Invention}
\begin{algorithm}[h]
\SetAlgoLined
\DontPrintSemicolon
\scriptsize
\caption{RePI: Reflective Problem Invention}
\KwIn{Failed problem $(I, q, a^*)$, Solver policy $\pi_S$, Teacher policy $\pi_T$}
\KwOut{New validated problem $(I', q', a')$}

\tcp{Step 1: Diagnose Solver Failure}
$\{a_S^{(i)}\}_{i=1}^k \leftarrow \pi_S(I, q)$ \tcp{Generate $k$ solution attempts}
$C \leftarrow \text{AnalyzeErrors}(\{a_S^{(i)}\}, a^*)$ \tcp{Identify reasoning gaps}

\tcp{Step 2: Invent New Problem}
$c \leftarrow \text{GetGeometryCode}(I, q)$ \tcp{Extract underlying code}
$c' \leftarrow \pi_T.\text{Modify}(c, C)$ \tcp{Modify code to address gaps}
$(I', q', a') \leftarrow \text{Execute}(c')$ \tcp{Render new problem}

\tcp{Step 3: Validate}
\If{$\text{GeometricValidation}(I', q') = \text{True}$}{
    \Return $(I', q', a')$
}
\Else{
    \Return $\text{RePI}(I, q, a^*)$ \tcp{Retry with different modification}
}
\end{algorithm}

\subsection{Reflect: Self-Verification}
\begin{algorithm}[h]
\SetAlgoLined
\DontPrintSemicolon
\scriptsize
\caption{Reflect: Self-Verification Module}
\KwIn{Candidate problem $(I', q', a')$, Teacher policy $\pi_T$}
\KwOut{Boolean validity flag}

\tcp{Step 1: Teacher Solves Problem}
$a_T \leftarrow \pi_T(I', q')$ \tcp{Teacher generates solution}

\tcp{Step 2: Compare with Reference}
$\text{match} \leftarrow \text{Compare}(a_T, a')$

\tcp{Step 3: Geometric Consistency Check}
$\text{valid} \leftarrow \text{CheckConstraints}(I', q')$

\Return $\text{match} \land \text{valid}$
\end{algorithm}

\section{Prompt Templates}
\label{app:prompts}

\subsection{Teacher Error Analysis Prompt}

\noindent\fbox{\parbox{0.95\linewidth}{
\small
\textbf{Error Analysis Prompt}

You are analyzing a geometry problem where the Solver failed. Given:
\begin{itemize}[leftmargin=12pt, itemsep=1pt]
    \item \textbf{Problem}: [Image + Question]
    \item \textbf{Reference Answer}: [Ground truth]
    \item \textbf{Solver Attempts}: [List of failed solutions]
\end{itemize}

Identify the specific reasoning gaps:
1. What geometric properties did the Solver miss?
2. What auxiliary constructions could help?
3. What constraints were violated?

Output a structured diagnosis for problem modification.
}}

\subsection{Teacher Problem Invention Prompt}

\noindent\fbox{\parbox{0.95\linewidth}{
\small
\textbf{Problem Invention Prompt}

Based on the error diagnosis, modify the geometry code to create a new problem that:
\begin{itemize}[leftmargin=12pt, itemsep=1pt]
    \item Directly addresses the identified reasoning gap
    \item Maintains geometric validity and non-degeneracy
    \item Produces a solvable problem with a unique answer
\end{itemize}

Modification strategies:
1. Add auxiliary lines/points to expose hidden relationships
2. Modify angle/length values to create new constraint patterns
3. Introduce additional geometric elements (circles, tangents, etc.)

Output executable Python geometry code.
}}

\subsection{Drawing Instruction Generation Prompt}

\noindent\fbox{\parbox{0.95\linewidth}{
\small
\textbf{Drawing Instruction Prompt}

Convert the following geometry problem into explicit drawing instructions for an image generator:

\textbf{Problem}: [Question text]
\textbf{Geometry Code}: [Python code]

Generate step-by-step drawing commands:
1. Specify exact coordinates and dimensions
2. Label all points, angles, and segments
3. Include visual styling (line thickness, colors)
4. Ensure mathematical precision

Output structured drawing instructions.
}}

\section{Supplementary Material Details}

\section{Training Curriculum Configuration}
\label{app:curriculum}

\textbf{Data Seed Construction}: The initial 108 seed problems comprise geometry word problems with accompanying figures and their corresponding Python code for geometric construction. These problems were manually selected from middle school exercise collections and exam papers by human experts, with complete annotations including problem statements, figures, and solution code.

\textbf{Geo170k Sampling Strategy}: From the Geo170k dataset, we selected 10k problems for training. Since each geometric figure in Geo170k corresponds to 11 problems with high homogeneity, we selected one representative problem per figure to ensure data quality and diversity while also maintaining a comparable quantity to other datasets.

\textbf{Stage 1}: Approximately 0.4k problems, consisting of: (1) 108 manually curated seed problems, and (2) augmented problems generated by the Teacher based on Solver errors during 8-attempt solving. All augmented problems are validated for geometric correctness with ground-truth solutions.

\textbf{Stage 2}: Approximately 1.0k problems, consisting of: (1) all 0.4k problems from Stage 1, and (2) augmented problems generated from Solver errors during 8-attempt solving of Stage 1. All augmented problems are validated with ground-truth solutions.

\textbf{Stage 3}: Approximately 2.5k problems, consisting of: (1) all 1.0k problems from Stage 2, and (2) augmented problems generated from Solver errors during 8-attempt solving of Stage 2. All augmented problems are validated with ground-truth solutions.

\section{Evaluation Protocol}
\label{app:evaluation}

\textbf{Answer Extraction Strategy}: We employ a two-tier extraction approach to identify student model responses:
\begin{itemize}
    \item \textbf{Primary Method}: Extract answers enclosed in \verb|\boxed{}| format using regex pattern \verb|\boxed\{([^}]+)\}|
    \item \textbf{Fallback Method}: If no boxed format detected, perform full-text search for option letters (A/B/C/D) or numerical values matching the ground truth
\end{itemize}

\textbf{Semantic Verification Module}: An LLM-based judge (Qwen3-VL-235B) evaluates answer correctness through structured comparison:
\begin{itemize}
    \item \textbf{Matching Rules}: Case-insensitive for multiple-choice options; unit-agnostic for numerical answers (e.g., ``90'' matches ``90 degrees'')
    \item \textbf{Temperature}: 0.1 for deterministic judgments
    \item \textbf{Max Tokens}: 10
    \item \textbf{Output Format}: Binary classification (``Correct'' or ``Incorrect'')
\end{itemize}

\textbf{Sampling Configuration}: 
\begin{itemize}
    \item \textbf{Prompting Strategy}: Zero-shot with three randomized prompt templates to reduce variance
    \item \textbf{Student Model Temperature}: 0.1
    \item \textbf{Max Tokens}: 1024
    \item \textbf{Repetition}: Each problem solved $N$ times independently (e.g., $N=8$ for Mean@8 metric)
    \item \textbf{Timeout}: 300s for student model, 120s for teacher model
\end{itemize}

\textbf{Mean@N Metric}: A problem is considered \textit{passed} if at least one attempt out of $N$ repetitions is judged correct by the teacher model. The final score is computed as:
$$\text{Mean@N} = \frac{\text{Number of Passed Problems}}{\text{Total Problems}}$$

\textbf{Implementation Details}: Evaluations run with configurable parallelism (default: 16 concurrent workers) using ThreadPoolExecutor. Trajectory files containing all intermediate results are saved incrementally every 10 problems to ensure fault tolerance and enable result inspection.

\section{Baseline Models}
\label{app:models}

We compared against the following representative models from GenExam's main results:

\textbf{Closed-source models}: GPT-Image-1~\citep{openai2025gptimage1}, Gemini-2.5-Flash-Image~\citep{google2025gemini}, Imagen-4-Ultra~\citep{deepmind2025imagen}, Seedream 4.0~\citep{bytedance2025seedream4}, Seedream 3.0~\citep{gao2025seedream3}, FLUX.1 Kontext max~\citep{batifol2025flux}.

\textbf{Open-source T2I models}: Qwen-Image~\citep{wu2025qwenimage}, HiDream-I1-Full~\citep{cai2025hidream}, FLUX.1 dev~\citep{batifol2025flux}, FLUX.1 Krea~\citep{batifol2025flux}, Stable Diffusion 3.5 Large~\citep{podell2024sd3}.

\textbf{Open-source unified MLLMs}: Show-o2~\citep{xie2025showo2}, BAGEL and BAGEL~(thinking)~\citep{li2024bagel}, Janus-Pro~\citep{chen2025januspro}, Emu3~\citep{wang2024emu3}, BLIP3o~\citep{chen2025blip3o}, BLIP3o-NEXT~\citep{chen2025blip3onext}.

\end{document}